\documentclass{article}

\PassOptionsToPackage{numbers, compress}{natbib}

\usepackage[final]{neurips_2022}

\usepackage[utf8]{inputenc} 
\usepackage[T1]{fontenc}    
\usepackage[colorlinks, linkcolor=red, citecolor=green]{hyperref} 
\usepackage{url}            
\usepackage{booktabs}       
\usepackage{amsfonts}       
\usepackage{nicefrac}       
\usepackage{microtype}      
\usepackage{xcolor}         

\usepackage{graphicx}
\usepackage{amsmath}
\newcommand{\ie}{\textit{i.e.}}
\newcommand{\eg}{\textit{e.g.}}
\usepackage{caption}
\usepackage{subcaption}

\usepackage[capitalize]{cleveref}
\crefname{section}{Sec.}{Secs.}
\Crefname{section}{Section}{Sections}
\Crefname{table}{Table}{Tables}
\crefname{table}{Tab.}{Tabs.}

\title{ModelScope Text-to-Video Technical Report}

\author{
    Jiuniu Wang\thanks{Equal contribution.} \hspace{0.2cm} Hangjie Yuan\footnotemark[1] \hspace{0.2cm} Dayou Chen\footnotemark[1]  \hspace{0.2cm} Yingya Zhang\footnotemark[1]\hspace{2pt} \thanks{Corresponding author.} \\ \hspace{0.2cm}  \textbf{Xiang Wang} \hspace{0.2cm}  \textbf{Shiwei Zhang}   \\ \\
    Alibaba Group
}

\begin{document}

\maketitle

\begin{abstract}
  This paper introduces ModelScopeT2V, a text-to-video synthesis model that evolves from a text-to-image synthesis model (\ie, Stable Diffusion).
  ModelScopeT2V incorporates spatio-temporal blocks to ensure consistent frame generation and smooth movement transitions.
  The model could adapt to varying frame numbers during training and inference, rendering it suitable for both image-text and video-text datasets.
  ModelScopeT2V brings together three components (\ie, VQGAN, a text encoder, and a denoising UNet), totally comprising 1.7 billion parameters, in which 0.5 billion parameters are dedicated to temporal capabilities.
  The model demonstrates superior performance over state-of-the-art methods across three evaluation metrics.
  The code and an online demo are available at \url{https://modelscope.cn/models/damo/text-to-video-synthesis/summary}.
\end{abstract}


\section{Introduction}



 

Artificial intelligence has expanded the boundaries of content generation in diverse modalities following simple and intuitive instructions.
This encompasses textual content~\cite{openai2023gpt4,brown2020gpt3,touvron2023llama}, visual content~\cite{nichol2021glide,huang2023composer,saharia2022Imagen,ramesh2022Dalle-2,kawar2022Imagic,zhou2022magicvideo,ho2022imagenvideo,luo2023videofusion,singer2022make-a-video,wang2023videocomposer} and auditory content~\cite{borsos2022audiolm,levkovitch2022zero-shot_voice,kong2021diffwave}.
In the realm of visual content generation, research efforts have been put into image generation~\cite{nichol2021glide,huang2023composer,saharia2022Imagen,ramesh2022Dalle-2} and editing~\cite{kawar2022Imagic}, leveraging diffusion models~\cite{sohl2015Diffusion_model}.


While video generation~\cite{skorokhodov2022stylegan-v,hong2022cogvideo,yu2022video_implicit_GAN} continues to pose challenges.
A primary hurdle lies in the training difficulty, which often leads to generated videos exhibiting sub-optimal fidelity and motion discontinuity. 
This presents ample opportunities for further advancements.
The open-source image generation methods (\eg, Stable Diffusion~\cite{rombach2022LDM}) have significantly advanced research in the text-to-image synthesis.
Nonetheless, the field of video generation has yet to benefit from a publicly available codebase, which could potentially catalyze further research efforts and progress.

To this end, we propose a simple yet easily trainable baseline for video generation, termed ModelScope Text-to-Video (ModelScopeT2V).
This model, which has been publicly available, presents two technical contributions to the field. 
Firstly, regarding the architecture of ModelScopeT2V, we explore  LDM~\cite{rombach2022LDM} in the field of text-to-video generation by introducing the spatio-temporal block that models temporal dependencies.
Secondly, regarding the pre-training technique, we propose a multi-frame training strategy that utilizes both the image-text and video-text paired datasets, thereby enhancing the model's semantic richness.
Experiments have shown that videos generated by ModelScopeT2V perform quantitatively and qualitatively similar or superior to other state-of-the-art methods.
We anticipate that ModelScopeT2V can serve as a powerful and effective baseline for future research related to the video synthesis, and propel innovative advancements and exploration. 

\begin{figure}[htbp]
    \centering
    \includegraphics[height=6cm]{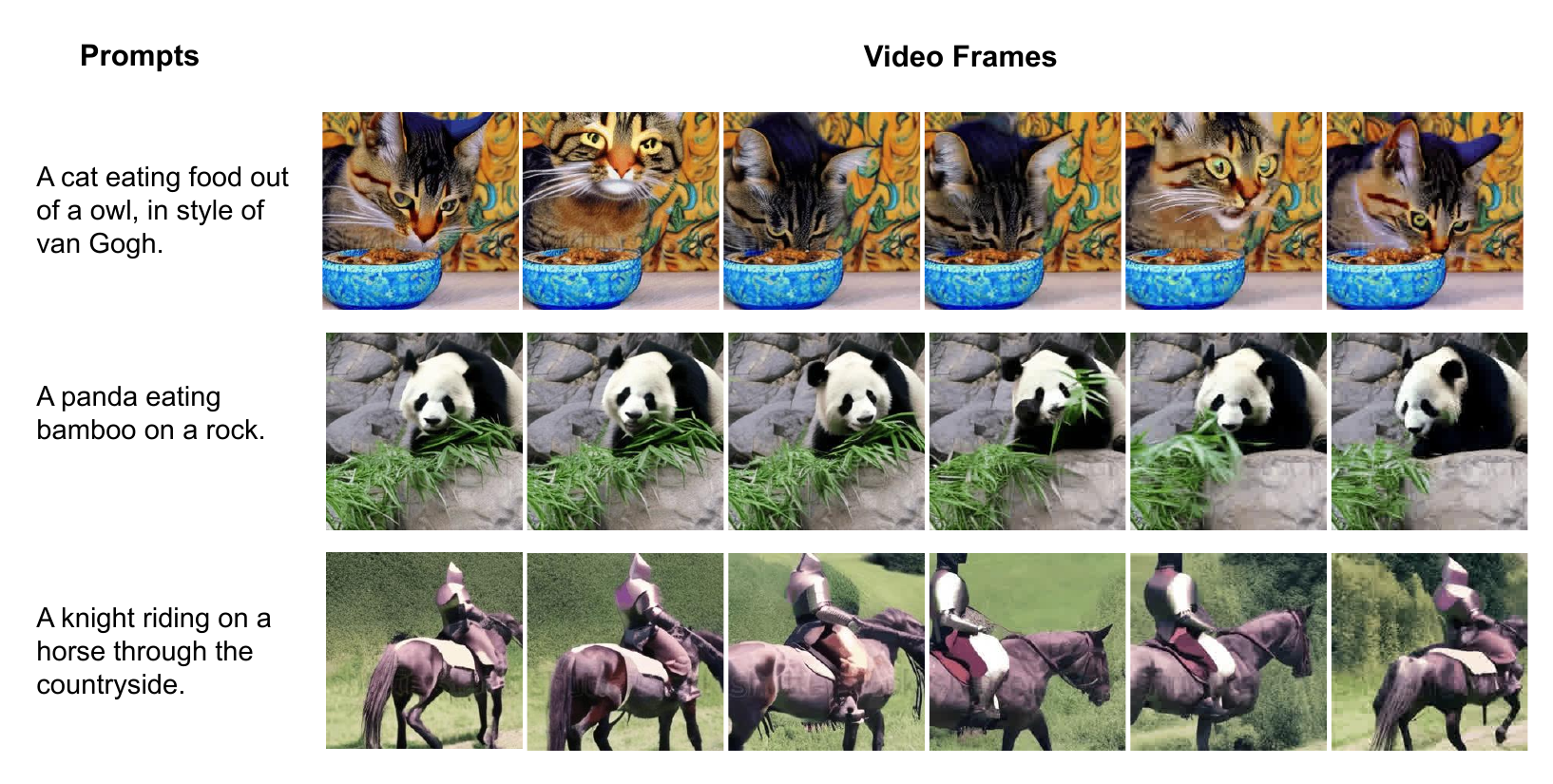}
    \caption{
    \textbf{The qualitative results of our ModelScopeT2V model.}
    The text-to-video-synthesis model could generate video frames adhering to the given prompts. 
    }
    \label{fig:teaser_fig}
\end{figure}
\section{Related work}

\textbf{Diffusion probabilistic models.}
Diffusion Probabilistic Models (DPM) was originally proposed in~\cite{sohl2015Diffusion_model}.
The early efforts of utilizing DPM into the image synthesis task at scale has been proven effective~\cite{kingma2021variational_diffusion_model,dhariwal2021diffusion_beat_gan}, surpassing dominant generative models, \eg, generative adversarial networks~\cite{goodfellow2020GAN} and variational autoencoders~\cite{kingma2013VAE}, in terms of diversity and fidelity.
The original DPM suffers the problem of low-efficiency when adopted for image/video generation due to the iterative denoising process and the high-resolution pixel space.
To solve the first obstacle, research efforts focused on improving the sampling efficiency by learning-free sampling~\cite{song2020denoising,song2020score_generative_SDE,liu2022pseudo_numerical,zhang2022fast_sampling_diffusion,lu2022dpm,zhang2022fast_sampling_diffusion} and learning based sampling~\cite{watson2022learning_fast_sampling,zheng2022truncated_diffusion,salimans2022progressive_distill_sampling}.
To address the second obstacle, methods like LDM~\cite{rombach2022high}, LSGM~\cite{song2020score_generative_SDE} and RDM~\cite{huang2022riemannian_diffusion_model} resorted to manifolds with lower intrinsic dimensionality~\cite{fefferman2016testing_manifold_hypo,yang2022diffusion_model_survey}. Our modelScopeT2V follows LDM~\cite{rombach2022high} but modifies it to the video generation task.

\textbf{Text-to-image synthesis via diffusion models.}
By receiving knowledge from natural language instructions (\eg, CLIP~\cite{radford2021CLIP} and T5~\cite{raffel2020T5}), diffusion models can be utilized for text-to-image synthesis.
LDM~\cite{rombach2022high} designed language-conditioned image generator by augmenting the UNet backbone~\cite{ronneberger2015UNet} with cross-attention layers~\cite{vaswani2017Transformer}.
DALL-E 2~\cite{ramesh2022Dalle-2} generated image embeddings for a diffusion decoder with CLIP text encoder.
The concurrent work, Imagen~\cite{saharia2022Imagen}, found the scalibility of T5, which means increasing the size of T5 could boost image fidelity and language-image alignment.
Building on existing image generation framework~\cite{rombach2022high,saharia2022Imagen}, Imagic~\cite{kawar2022Imagic} achieves text-based semantic image edits by leveraging intermediate text embeddings that align with the input image and the target text.
Composer~\cite{huang2023composer} reformulates images as various compositions, thus enabling image generation from not just texts, but also sketches, masks, depthmaps and more.
The ModelScopeT2V initialize the spatial part from Stable Diffusion model~\cite{rombach2022high}, and proposes the spatio-temporal block that empowers the capacity of temporal dependencies.

\textbf{Text-to-video synthesis via diffusion models.}
Generating realistic videos remains challenging due to the difficulty in generating videos with high fidelity and motion continuity~\cite{skorokhodov2022stylegan-v,hong2022cogvideo,yu2022video_implicit_GAN}.
Recent works have utilized diffusion models~\cite{sohl2015Diffusion_model,lu2022dpm} to generate authentic videos~\cite{yang2022DPM_video,harvey2022flexible_diffusion_video,hoppe2022diffusion_video_infilling,wang2023videocomposer}.
Text, as a highly intuitive and informative instruction, has been employed to guide video generation.
Various approaches have been proposed, such as Video Diffusion~\cite{ho2022video_diffusion_models}, which introduces a spatio-temporal factorized 3D Unet~\cite{cciccek20163dUNet} with a novel conditional sampling mechanism.
Imagen Video~\cite{ho2022imagenvideo} synthesizes high definition videos given a text prompt by designing a video generator and a video super-resolution model.
Make-A-Video~\cite{singer2022make-a-video} employs off-the-shelf text-to-image generation model combined with spatio-temporal factorized diffusion models to generate high-quality videos without relying on paired video-text data.
Instead of modeling the video distribbution in the visual space (\eg, RGB format), MagicVideo~\cite{zhou2022magicvideo} designed a generator in the latent space with a distribution adapter, which did not require temporal convolution. 
In oder to improve the content and motion performance, VideoFusion~\cite{luo2023videofusion} decouples per-frame noise into base noise and residual noise, which benefits from a well-pretrained DALL-E 2 and provides better control over content and motion.
Targetting decoupling, Gen-1~\cite{esser2023gen-1} defines the content latents as CLIP embeddings and the structure latents as the monocular depth estimates, attaining superior decoupled controllability.
ModelScopeT2V proposes a simple yet effective training pipeline that benefits from semantic diversity inherent in image-text and video-text paired datasets, further enhancing the learning process and performance in video generation.








\section{Methodology}
In this section, we introduce the overall architecture of ModelScopeT2V (\cref{sec:latent_vid_diff_model}), the key spatio-temporal block used in UNet (\cref{sec:spatiotemporal_block}) and a multi-frame training mechanism that stabilizes training (\cref{sec:multi-frame_training}).

\begin{figure*}
    \centering
    \includegraphics[height=6cm]{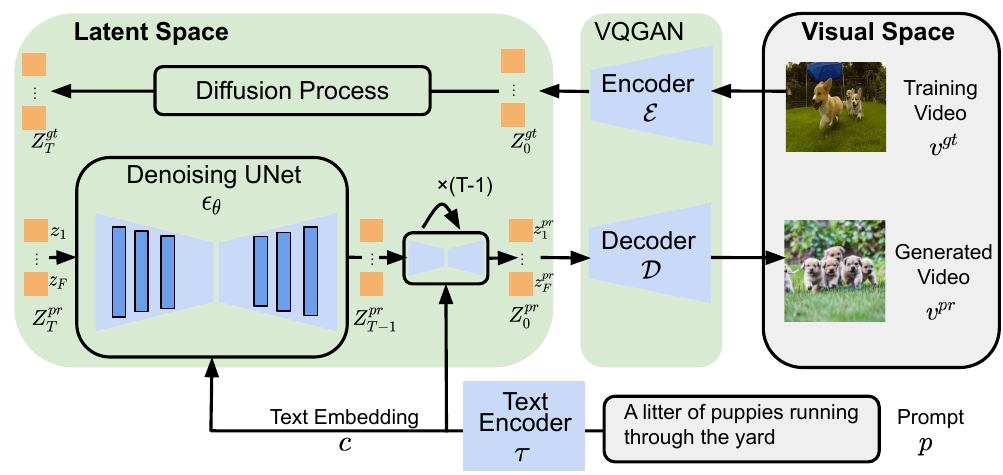}
    \caption{
    \textbf{The overall architecture of ModelScopeT2V.} 
    Here, a text encoder $\tau$ encodes the  prompt $p$ into text embedding $c$.
    Then, input the embedding $c$ into the UNet$\epsilon_{\theta}$, directing the denoising process.
    During training, a diffusion process is performed, transitioning from $Z^{gt}_0$ to $Z^{gt}_T$; so the denoising UNet could be trained on these latent variables.
    Conversely, during inference, random noise $Z^{pr}_0$ is sampled and utilized for the denoising procedure.
    }
    \label{fig:main_structure}
\end{figure*}

\subsection{ModelScopeT2V}
\label{sec:latent_vid_diff_model} 

\textbf{Structure overview.} Given a text prompt $p$, ModelScopeT2V outputs a video $v^{pr}$ through a latent video diffusion model that conforms to the semantic meaning of the prompt.
The architecture of the latent video diffusion model is shown in~\cref{fig:main_structure}. 
As illustrated in the figure, the training video $v^{gt}$ and generated video $v^{pr}$ are in the visual space. 
The diffusion process and denoising UNet $\epsilon_{\theta}$ are in the latent space. 
VQGAN~\cite{PatrickEsser2021TamingTF} converts the data between visual space and latent space through its encoder $\mathcal{E}$ and decoder $\mathcal{D}$. 
The latent space in this paper is proposed by VQGAN~\cite{PatrickEsser2021TamingTF}. Given a training video $v^{gt} = [f_1, \dots, f_F]$ with $F$ frames, we could encode the video with VQGAN encoder $\mathcal{E}$ as,
\begin{align}
	Z^{gt}_0 = [\mathcal{E}(f_1), \dots, \mathcal{E}(f_F)] \,, \label{equ:encoder}
\end{align}
where $v^{gt} \in \mathbb{R}^{F \times H \times W \times 3}$ is a RGB video, $Z^{gt}_0 \in \mathbb{R}^{F \times \frac{H}{8} \times \frac{W}{8} \times 4}$ is the ground-truth latent variable, $H$ and $W$ are the height and width of frames.
The denoising UNet works on the latent space. During training, the diffusion process turns $Z^{gt}_0$ into $Z^{gt}_T$ by adding gaussian noise $[\epsilon^{gt}_{1}, \dots, \epsilon^{gt}_{T}]$ for $T$ steps. 
Therefore, we have $[Z^{gt}_0, \dots, Z^{gt}_T]$, which contains less information as the diffusion process proceeds.
During inference, the UNet predicts the added noise for each step, so that we finally generate $Z^{pr}_0=[z^{pr}_1,\dots,z^{pr}_F]$ from a random noise $Z^{pr}_T$. Then we could generate a video $v^{pr}$ by VQGAN decoder $\mathcal{D}$ as:
\begin{align}
    v^{pr}  = [\mathcal{D}(z^{pr}_1), \dots, \mathcal{D}(z^{pr}_F)] \,,
\end{align}

\paragraph{Text conditioning mechanism.} 
ModelScopeT2V aims to generate videos conforming to the given text prompts.
Therefore, it is desired to ensure the textual controllability by effectively injecting textual information into the generative process.
Inspired from Stable Diffusion~\cite{rombach2022high}, we augment the UNet structure with a cross-attention mechanism, which is an effective approach to condition the visual content on texts~\cite{Yuan2022RLIP,jaegle2021perceiver,huang2023composer,li2021GLIP}.
Specifically, we use the text embedding $c$ of the prompt $p$ in the spatial attention block as the key and value in the multi-head attention layer. 
This enables the intermediate UNet features to aggregate text features seamlessly, thereby facilitating an alignment of language and vision embeddings.
To ensure a great alignment between language and vision, we utilize the text encoder from pre-trained CLIP ViT-H/14~\cite{radford2021CLIP} to convert the prompt $p$ into the text embedding $c \in \mathbb{R}^{N_p \times N_c}$ where $N_p$ represents the maximum token length of the prompt, and $N_c$ represents the dimension of the token embedding.



\paragraph{Denoising UNet.}  The UNet includes different types of blocks such as the initial block, downsampling block, spatio-temporal block and upsampling block, represented by the dark blue squares in Figure~\ref{fig:main_structure}. Most parameters of the model are concentrated in the denoising UNet. So the denoising UNet $\epsilon_{\theta}$ is considered as the core of the latent video diffusion model, which performs the diffusion process in the latent space. 
The UNet aims to denoise from $Z_T$ to $Z_0$ by predicting the noise of each step. 
Given a specific step index $\hat{t} \in [1, 2,\dots,T]$, the predicted noise $\epsilon^{pr}_{\hat{t}}$ can be formulated as:
\begin{align}
        c &= \tau(p) \\
	\epsilon^{pr}_{\hat{t}}  &= \epsilon_{\theta}(Z_{\hat{t}},c,\hat{t})
\end{align}
where $p$ denotes the prompt, $c$ represent the text embedding, and $Z_{\hat{t}}$ is the latent variable in $\hat{t}$-th step. 
In this case, $Z_{\hat{t}}$ denotes $Z^{gt}_{\hat{t}}$ during training, and denotes the denoised latent video representation $Z^{pr}_{\hat{t}}$ during inference.
The model's objective is to minimize the discrepancy between the predicted noise $\epsilon^{pr}_{\hat{t}}$ and ground-truth noise $\epsilon^{gt}_{\hat{t}}$. 
Consequently, the training loss $\mathcal{L}$ of the UNet can be formulated as:
\begin{align}
        \mathcal{L} = \mathbb{E}_{Z_{\hat{t}},\epsilon^{gt}_{\hat{t}} \sim \mathcal{N}(0,1),\hat{t}} \left [ ||\epsilon^{gt}_{\hat{t}}  - \epsilon^{pr}_{\hat{t}} ||_2^2 \right ] 
\end{align}
which means we hope to decrease the mathematical expectation $\mathbb{E}$ of the difference between $\epsilon^{gt}_{\hat{t}}$ and $\epsilon^{pr}_{\hat{t}}$.

\subsection{Spatio-temporal block}
\label{sec:spatiotemporal_block}

Our video diffusion model is built upon a UNet architecture, which consists of four key building blocks: the initial block, the downsampling block, the spatio-temporal block and the upsampling block.
The initial block projects the input into the embedding space, while the downsampling and upsampling blocks spatially downsample and upsample the feature maps, respectively.
The spatio-temporal block plays a crucial role in capturing complex spatial and temporal dependencies in the latent space, thereby enhancing the quality of video synthesis.
To this end, we leverage the power of spatio-temporal convolutions and attentions to comprehensively obtain such complex dependencies.

\begin{figure}
        \centering
        \includegraphics[height=3.3cm]{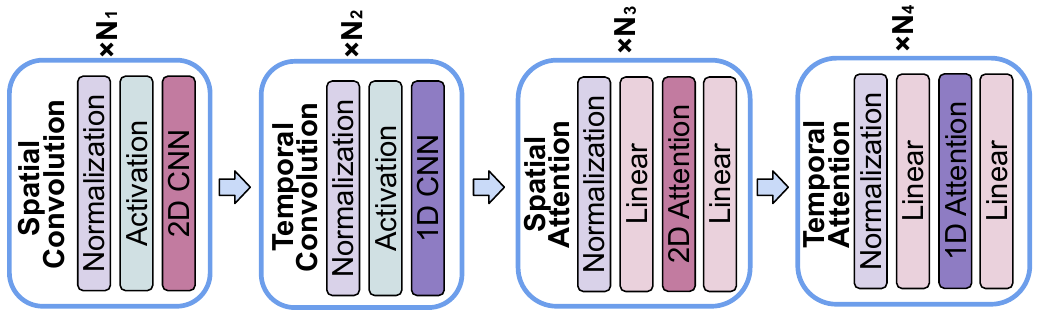}
        \caption{
        \textbf{The structure of the spatio-temporal block.} It includes four modules, \ie, spatial convolution, temporal convolution, spatial attention, and temporal attention. The main layers for these modules are marked in  different colors.
        }
        \label{fig:block_structure}
\end{figure}

\paragraph{Structure overview.}
Figure~\ref{fig:block_structure} illustrates the architecture of the spatio-temporal block in our video diffusion model.
To ensure effectively synthesising videos, we factorise the convolution and the attention mechanism over space and time~\cite{qiu2017P3D,arnab2021vivit}.
Therefore, the spatio-temporal block is composed of four sub-components, namely spatial convolution, temporal convolution, spatial attention and temporal attention.
The spatio-temporal convolutions capture correlations across frames by convolving over both the spatial and temporal dimensions of the video, while the spatio-temporal attentions capture correlations across frames by selectively attending to different regions and time steps within the video.
By utilizing these spatio-temporal building blocks, our model can effectively learn spatio-temporal representations and generate high-quality videos.
Specifically, one spatio-temporal block consists of $N_1$ spatial convolutions, $N_2$ temporal convolutions, $N_3$ spatial attention, and $N_4$ temporal attention operations.
In our experiments, we set $(N_1, N_2, N_3, N_4) = (2, 4, 2, 2)$ by default to achieve a balance between performance and computational efficiency.
Regarding the architecture of the two spatial attentions, we instantiate it in two different ways.
The first attention is a cross-attention module that conditions the visual features on the textual features, allowing cross-modal interactions.
The other attention is a self-attention module that operates solely on visual features, responsible for spatial modeling. While the two temporal attentions are both self-attention module.

\begin{figure}[!tb]
\centering
\subcaptionbox{Spatio-temporal Convolution. \label{fig:temporal_conv}}
{
\includegraphics[width=0.6\linewidth]{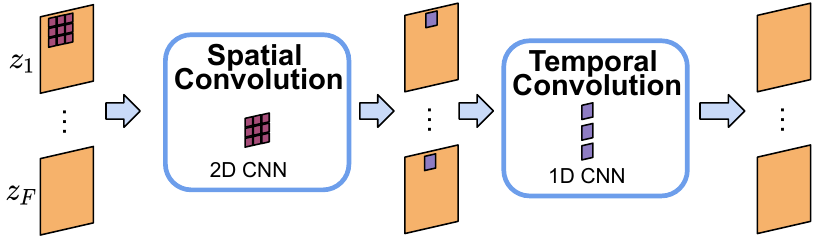}
}
\subcaptionbox{Spatio-temporal Attention.\label{fig:temporal_attention}}
{
\includegraphics[width=0.6\linewidth]{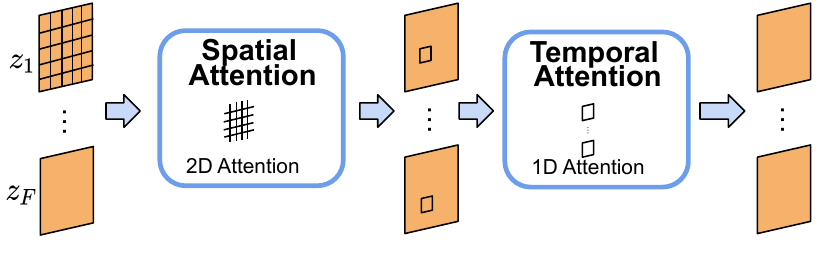}
}
\caption{
\textbf{Diagram of the processing flow for the spatio-temporal block.} Here $z_i$ denotes the latent variable of the $i$-th frame.
(a) displays the core structure of spatio-temporal convolution, including a 2D CNN with a kernel size of $3 \times 3$ and a 1D CNN with a kernel size of 3. (b) shows the variable processing in multi-head attention of spatio-temporal attention, the frame variable is flattened at the spatial scale in spatial attention; while variables at the same positions between frames are grouped together in temporal attention.\label{fig:spatio-temporal}}
\end{figure}

\paragraph{Spatio-temporal convolutions.} Figure~\ref{fig:temporal_conv} illustrates the spatio-temporal convolution, which is composed of both spatial and temporal convolutions. 
The spatial convolution employs a convolution kernel of size $3 \times 3$ to extract features from the $\frac{H}{8} \times \frac{W}{8}$ latent features within each frame, where $H$ and $W$ denote the height and width of video frames in visual space.
Meanwhile, the temporal convolution adopts a convolution kernel of size $3$ to extract features from $F$ frames, where $F$ represents the number of frames for each video.

\paragraph{Spatio-temporal attentions.} Figure~\ref{fig:temporal_attention} displays the spatio-temporal attention, which consists of spatial and temporal attention modules.
In detail, the spatial attention operates on the latent features in the spatial dimension of size $\frac{HW}{64} $, while the temporal attention operates on the temporal dimension of size $F$.
We adopt the popular Transformer architecture~\cite{vaswani2017Transformer} to instantiate both attention mechanisms.



\subsection{Multi-frame training} 
\label{sec:multi-frame_training}
ModelScopeT2V is designed to be trained on large-scale video-text paired datasets, such as WebVid~\cite{Bain21}, which is domain-aligned with video generation.
Nonetheless, the scale of such datasets is orders of magnitude smaller compared to image-text paired datasets, such as LAION~\cite{schuhmann2021laion}.
Despite initializing the spatial part of ModelScopeT2V with Stable Diffusion~\cite{rombach2022high}, training solely on video-text paired datasets can hinder semantic diversity and lead to catastrophic forgetting of image-domain expertise during training~\cite{li2017LwF,feng2022overcoming,kirkpatrick2017overcomingforgetting,Feng2022PLwF}.
To overcome this limitation and leverage the strengths of both datasets, we propose a multi-frame training approach.
Specifically, one eighth of GPUs for training are applied to image-text paired datasets, while the remaining GPUs handle video-text paired datasets.
Since the model structure could adapt to
 any frame length, one image could be considered as a video with frame length 1 for those GPUs training on image-text paired datasets.

\section{Experiments}

\subsection{Implementation details}

\subsubsection{Datasets}

\paragraph{LAION~\cite{schuhmann2021laion}.} We utilize the LAION-5B dataset as image-text pairs, specifically the LAION2B-en subset, as the model focuses on English input. The LAION dataset encompasses objects, people, scenes, and other real-world elements. 

\paragraph{WebVid~\cite{Bain21}.} The WebVid dataset comprises almost 10 million video-text pairs, with a majority of the videos having a resolution of $336 \times 596$. Each video clip has a duration of approximately 30 seconds. During the model training, we selected the middle square portion and randomly picked 16 frames with 3 frames per second as training data.

\paragraph{MSR-VTT~\cite{xu2016msr-vtt}.} The MSR-VTT dataset is used to validate our model performance and is not utilized for training. This dataset includes 10k video clips, each of which is annotated with 20 sentences. To obtain FID-vid and FVD metric results, 2,048 video clips were randomly selected from the test set, and one sentence was randomly chosen from each clip to generate videos. When evaluating on CLIPSIM metric, we followed previous works~\cite{singer2022make-a-video,wu2022nuwa} and use nearly 60k sentences from the whole test split as prompts to generate videos.

\subsubsection{Model instantiation and hyper-parameters}
We employ DDPM~\cite{ho2020denoising} with $T = 1,000$ steps for training and use DDIM sampler~\cite{song2020denoising} in classifier-free guidance~\cite{ho2022classifier} with $50$ steps for inference by default.
ModelScopeT2V primarily consists of three modules: the Text encoder $\tau$, VQGAN, and Denoising UNet. 
The pretrained checkpoint for initializing VQGAN and Denoising UNet are obtained from Stable Diffusion~\cite{rombach2022high} version 2.1\footnote{https://github.com/Stability-AI/stablediffusion}.
The parameters in VQGAN remains frozen during training and inference.
The outputs of temporal convolution and temporal attention are initialized as zeros, enabling ModelScopeT2V to generate meaningful yet temporally discontinuous frames at the beginning of training.
As the training progresses, the temporal structures will be optimized to learn the temporal correspondence between frames, thereby synthesising continuous videos.


\subsubsection{Training details}

We train ModelScopeT2V using the AdamW optimizer~\cite{loshchilov2017AdamW} with a learning rate of $5\times 10^{-5}$.
Our model is trained on 80G NVIDIA A100 GPUs.
We perform multi-frame training as detailed in Section~\ref{sec:multi-frame_training}, specifically using a batch size of 1,400 for images and a batch size of 3,200 for videos, and training 267 thousand iterations. 
The compression factor of VQGAN is 8, meaning that it converts RGB images of size $256 \times 256$ into latent representations of size $32 \times 32$ .
For the text encoder, we set the maximum text length to $N_p=77$, and embedding dim $N_c=768$, which are consistent with the pre-trained OpenCLIP\footnote{https://github.com/mlfoundations/open\_clip}.

We empirically observe that employing either temporal convolution or temporal attention can augment ModelScopeT2V's ability to capture temporal dependency.
This observation is partly supported by VideoCraft\footnote{https://github.com/VideoCrafter/VideoCrafter} which only contains temporal attention for temporal modeling.
We take a step further by employing both the temporal convolution and the temporal attention, which facilitates the ModelScopeT2V to achieve superior temporal modeling.
In detail, we use $N_2 = 4$ temporal convolution blocks and $N_4 = 2$ temporal attention block for each spatio-temporal block.
These temporal blocks account for 552 million parameters out of the total 1,345 million parameters in our UNet, indicating that 39\% parameters of the UNet parameters are dedicated to capturing temporal information. 
As a result, the entire ModelScopeT2V model (including VQGAN and the text encoder) comprises approximately 1.7 billion parameters.

We observe that use more layers of temporal convolution would lead to better temporal ability. Since the kernel size of 1D CNN in temporal convolution is 3, $N_2=4$ temporal convolution layers could lead the local receptive field as 81 in each spatio-temporal block, which is enough for 16 output frames per video. For multi-frame training, the temporal convolution and temporal attention mechanisms are still active. Our experiments show it is unnecessary to change the range of parameters for different frame settings.

\subsection{Main results}

\subsubsection{Quantitative results}
ModelScopeT2V is evalutated on MSR-VTT~\cite{xu2016msr-vtt} dataset. 
We conduct the evaluation under a zero-shot setting since ModelScopeT2V is not trained on MSR-VTT. 
We compare ModelScopeT2V with several state-of-the-art models using FID-vid~\cite{heusel2017gans_nash_equilibrium}, FVD~\cite{unterthiner2018FVD}, and CLIPSIM~\cite{wu2021godiva} metrics. 
The FID-vid and FVD are assessed based on 2,048 randomly selected videos from MSR-VTT test split, where we compute the metrics using the middle 16 frames of each video with an FPS of 3. 
CLIPSIM are evaluated based on all captions from MSR-VTT test split following~\cite{singer2022make-a-video}.
The resolution of the generated videos is consistently $256 \times 256$.

As shown in Table~\ref{tab:results}, ModelScopeT2V achieves the best performance on both FID-vid (\ie, 11.09) and FVD (\ie, 550), indicating that our generated videos are visually similar to the ground truth videos.
Our model also obtains a competitive score of 0.2930 on CLIPSIM, suggesting that our generated videos are semantically similar to the text prompts. 
The CLIPSIM score of our model is only marginally lower than that of Make-A-Video~\cite{singer2022make-a-video}, while they utilize additional data from HD-VILA-100M~\cite{xue2022HD-VILA-100M} for training.

\begin{table}[t]
\centering
\begin{tabular}{c|ccc}
\hline
\textbf{Models}              & FID-vid ($\downarrow$)   &  FVD ($\downarrow$) & CLIPSIM ($\uparrow$) \\ \hline
N\"UWA~\cite{wu2022nuwa}       & 47.68   & -  & 0.2439        \\
CogVideo (Chinese)~\cite{hong2022cogvideo}   & 24.78 & - & 0.2614      \\
CogVideo (English)~\cite{hong2022cogvideo}   & 23.59  & 1294 & 0.2631  \\
MagicVideo~\cite{zhou2022magicvideo}          & -    & 1290 & -        \\
Video LDM~\cite{blattmann2023align}      & -   &  - & 0.2929 \\ 
Make-A-Video~\cite{singer2022make-a-video}        & 13.17 &  -& \textbf{0.3049}     \\ \hline
ModelScopeT2V (ours) &   \textbf{11.09} & \textbf{550} & 0.2930    \\ \hline
\end{tabular}
\caption{
\textbf{Quantitative comparison with state-of-the-art models on MSR-VTT.} We evaluate the models with three metrics (\ie, FID-vid~\cite{heusel2017gans_nash_equilibrium}, FVD~\cite{unterthiner2018FVD}, and CLIPSIM~\cite{wu2021godiva}).
}
\label{tab:results}
\end{table}

\subsection{Qualitative results} 
In this subsection, we compare the qualitative results of ModelScopeT2V with other state-of-the-art methods. 
To facilitate comparison with Make-A-Video and Imagen Video, generated video frames with the same frame index are presented in the same column with ModelScopeT2V.
The videos generated by Make-A-Video~\cite{singer2022make-a-video}\footnote{https://makeavideo.studio} and Imagen Video~\cite{ho2022imagenvideo}\footnote{https://imagen.research.google/video} were downloaded from their official webpages.
Six frames are uniformly sampled from each video for comparison. 
This comparison is fair in terms of video duration, as all three methods (\ie, Make-A-Video, Imagen Video, and ModelScopeT2V) generate 16-frame videos aligned with given texts.
One difference is that Imagen Video generates videos with a aspect ratio of $2:5$, while the other two generate videos with a aspect ratio of $1:1$.

\begin{figure}[htb]
        \centering
        \includegraphics[height=15cm]{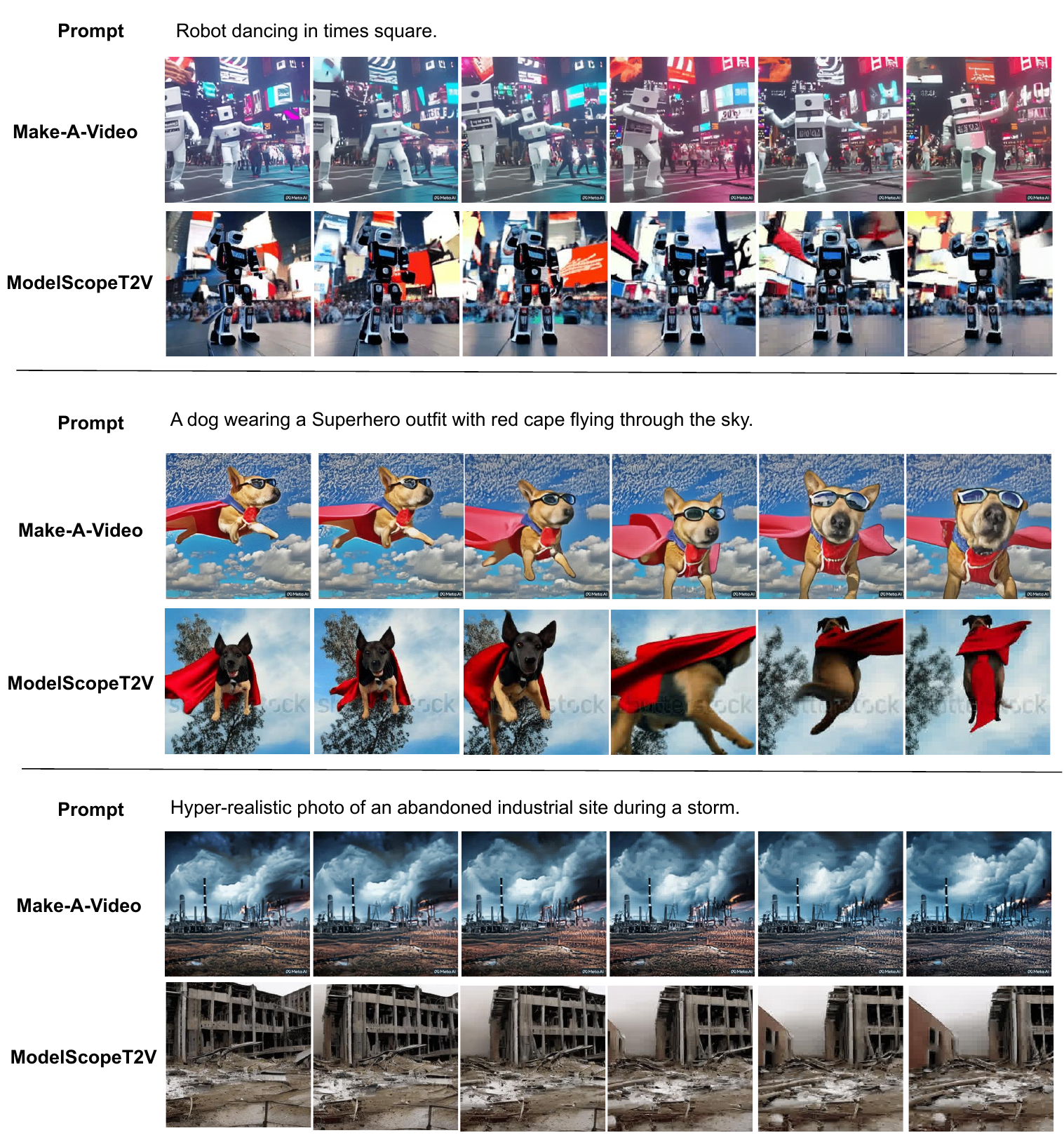}
        \caption{
        \textbf{Qualitative results comparison with Make-A-Video.}
        We present three examples, each displaying the provided prompt, the video generated by Make-A-Video and the video generated by ModelScopeT2V.
        }
        \label{fig:compare_with_make_a_video}
\end{figure}

The quanlitative comparison between ModelScopeT2V and Make-A-Video is displayed in Figure~\ref{fig:compare_with_make_a_video}.
We can observe that both methods generate videos of high quality, which is consistent with the quantitative results. 
However, the ``robot'' in the first example and the ``dog'' in the second example generated by ModelScopeT2V exhibit a superior degree of realism.
We attribute this advantage to our model's joint training with image-text pairs, which enhances its comprehension of the correspondence between textual and visual data.
In the third example, while the ``industrial site'' generated by Make-A-Video is more closely aligned with the prompt, depicting the overall scene of a ``storm'', ModelScopeT2V produces a distinctive interpretation showcasing two ``abandoned'' factories and a gray sky, captured from various camera angles.
This difference stems from Make-A-Video's use of image CLIP embedding to generate videos, which can result in less dynamic motion information.
In general, ModelScopeT2V demonstrates a wider range of motion in its generated videos, distinguishing it from Make-A-Video.



\begin{figure}[htb]
        \centering
        \includegraphics[height=13.5cm]{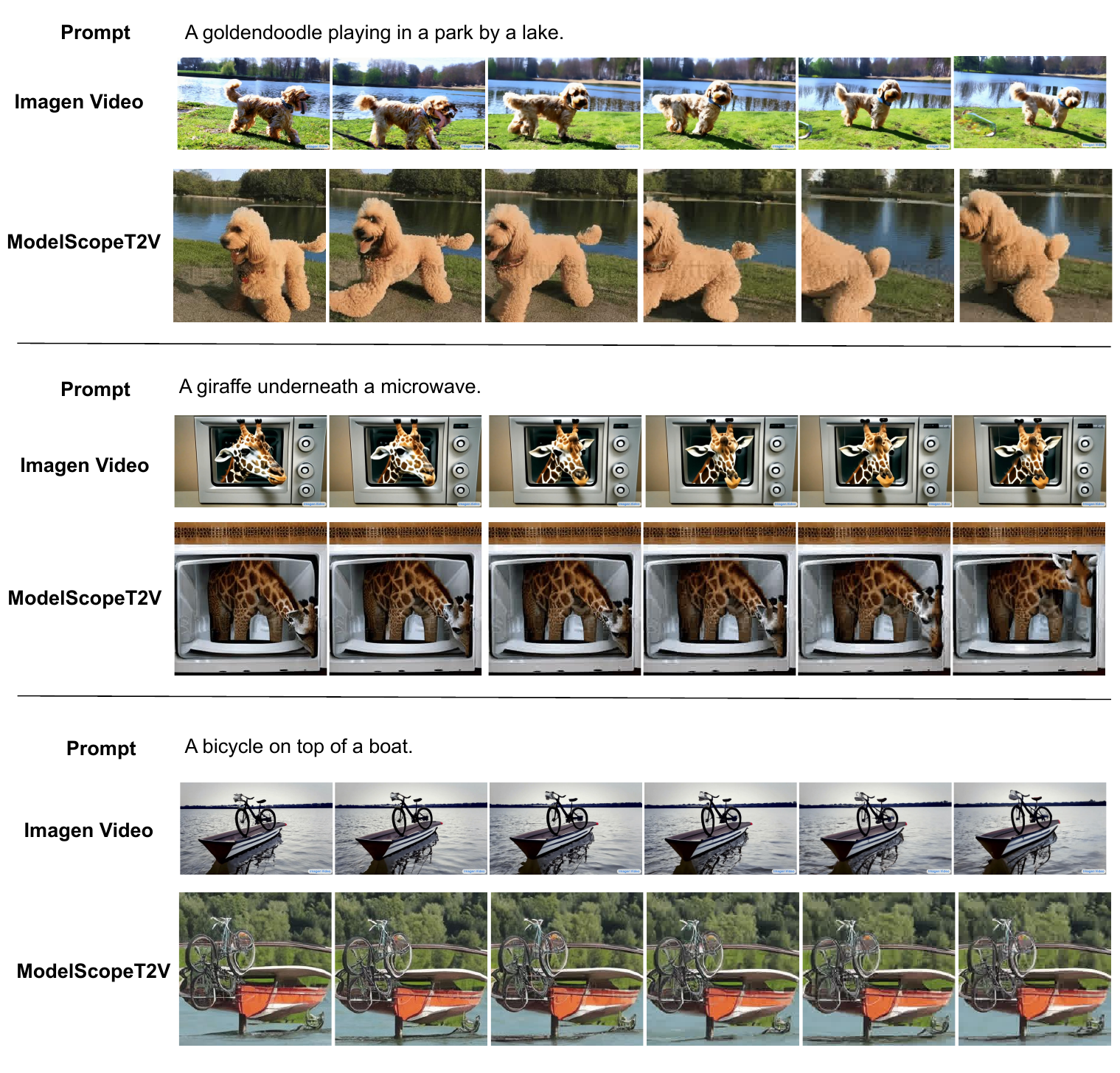}
        \caption{
        \textbf{Qualitative results comparison with Imagen Video.}
        We present three examples, each displaying the provided prompt, the video generated by Imagen Video and the video generated by ModelScopeT2V.
        }
        \label{fig:compare_with_imagen}
\end{figure}


The comparison of our method, ModelScopeT2V, with Imagen Video is illustrated in Figure~\ref{fig:compare_with_imagen}.
While Imagen Video generate more vivid and contextually relevant video content, ModelScopeT2V effectively depicts the content of the prompt, albeit with some roughness in the details.
For instance, in the first example of Figure~\ref{fig:compare_with_imagen}, it's noteworthy that Imagen Video generates a video whose second frame illustrates a significantly distorted dog's tongue, exposing the model's limitations in accurately rendering the real world.
On the other hand, ModelScopeT2V demonstrates its potential in robustly representing the content described in the prompt.
It's worth noting that the superior performance of Imagen Video can be attributed to the employment of the T5 text encoder, a base model with a larger number of parameters, and a larger-scale training dataset, which is not utilized in ModelScopeT2V.
Considering the performance, our  ModelScopeT2V lays a strong foundation for future improvements and shows considerable promise in the domain of text-to-video generation.


\subsection{Community development} 

We have made the code for ModelScopeT2V publicly available on the GitHub repositories of ModelScope\footnote{https://github.com/modelscope/modelscope/blob/master/modelscope/models/multi\_modal/video\_synthesis} and Diffuser\footnote{https://huggingface.co/spaces/damo-vilab/modelscope-text-to-video-synthesis/blob/main/app.py}.
Additionally, we have provided online demos of ModelScopeT2V on ModelScope\footnote{https://modelscope.cn/studios/damo/text-to-video-synthesis/summary} and HuggingFace\footnote{ https://huggingface.co/spaces/damo-vilab/modelscope-text-to-video-synthesis}. 
The open-source community has actively engaged with our model and uncovered several applications of ModelScopeT2V.
Notably, projects such as sd-webui-text2video\footnote{https://github.com/deforum-art/sd-webui-text2video} and Text-To-Video-Finetuning\footnote{https://github.com/ExponentialML/Text-To-Video-Finetuning}have extended the model's usage and broadened its applicability.
Additionally, the video generation feature of ModelScopeT2V has already been successfully utilized for the creation of short videos\footnote{https://youtu.be/Ank49I99EI8}.

\section{Conclusion}
This paper proposes ModelScopeT2V, the first open-source diffusion-based text-to-video generation model.
To enhance the ModelScopeT2V's ability to modeling temporal dynamics, we design the spatio-temporal block that incorporates spatio-temporal convolution and spatio-temporal attention.
Furthermore, to leverage semantics from comprehensive visual content-text pairs, we  perform multi-frame training on both text-image pairs and text-video pairs.
Comparative analysis of videos generated by ModelScopeT2V and those produced by other state-of-the-art methods demonstrate similar or superior performance quantitatively and qualitatively.

As for future research directions, we expect to adopt additional conditions to enhance video generation quality.
Potential strategies include using multi-condition approaches~\cite{lhhuang2023composer} or the LoRA technique~\cite{hu2021lora}.
Additionally, an interesting topic to explore could be the generation of longer videos that encapsulate more semantic information.

\bibliographystyle{plain}
\bibliography{refs}

\end{document}